\documentclass[10pt,twocolumn,letterpaper]{article}

\usepackage{cvpr}
\usepackage{times}
\usepackage{epsfig}
\usepackage{graphicx}
\usepackage{amsmath}
\usepackage{amssymb}

\usepackage[pagebackref=true,breaklinks=true,letterpaper=true,colorlinks,citecolor=green,linkcolor=red,bookmarks=false]{hyperref}

\usepackage{multirow}
\usepackage{authblk}

\usepackage[breaklinks=true,bookmarks=false]{hyperref}

\cvprfinalcopy % *** Uncomment this line for the final submission

\def\cvprPaperID{****} % *** Enter the CVPR Paper ID here

\begin{document}

%%%%%%%%% TITLE
\title{Three-Stream Convolutional Networks for Video-based \\ Person Re-Identification}

%\author[1]{Author A\thanks{A.A@university.edu}}
%\author[1]{Author B\thanks{B.B@university.edu}}
%\author[1]{Author C\thanks{C.C@university.edu}}
%\author[2]{Author D\thanks{D.D@university.edu}}
%\author[2]{Author E\thanks{E.E@university.edu}}
%\affil[1]{Department of Computer Science }
%\affil[2]{Department of Mechanical Engineering}

\author[1,3]{Zeng Yu}
\author[1]{Tianrui Li}
\author[2]{Ning Yu}
\author[1]{Xun Gong}
\author[1]{Ke Chen}
\author[3]{Yi Pan}
\affil[1]{School of Information Science and Technology, Southwest Jiaotong University}
\affil[2]{Department of Computing Sciences, The College at Brockport State University of New York}
\affil[3]{Department of Computer Science, Georgia State University}

%
%\author{Zeng Yu\\
%Southwest Jiaotong University\\
%%Institution1 address\\
%{\tt\small zyu7@gsu.edu}
%% omit the following lines up until the closing ``}''.
%% Additional authors and addresses can be added with ``\and'',
%% just like the second author.
%% To save space, use either the email address or home page, not both
%\and
%Tianrui Li\\
%Southwest Jiaotong University\\
%%First line of institution2 address\\
%{\tt\small trli@swjtu.edu.cn}
%}
%
%
%%\and
%%Xiuchun Xiao\\
%%Southwest Jiaotong University\\
%%%First line of institution2 address\\
%%{\tt\small trli@swjtu.edu.cn}
%%\and
%%Shiwei Zhen\\
%%Southwest Jiaotong University\\
%%%First line of institution2 address\\
%%{\tt\small trli@swjtu.edu.cn}
%%\and
%%Ke Chen\\
%%Southwest Jiaotong University\\
%%%First line of institution2 address\\
%%{\tt\small trli@swjtu.edu.cn}
%%}

%% omit the following lines up until the closing ``}''.
%% Additional authors and addresses can be added with ``\and'',
%% just like the second author.
%% To save space, use either the email address or home page, not both
%\and

\maketitle
%\thispagestyle{empty}

%%%%%%%%% ABSTRACT
\begin{abstract}

This paper aims to develop a new architecture that can make full use of the feature maps of convolutional networks. To this end, we study a number of methods for video-based person re-identification and make the following findings: 1) Max-pooling only focuses on the maximum value of a receptive field, wasting a lot of information. 2) Networks with different streams even including the one with the worst performance work better than networks with same streams, where each one has the best performance alone. 3) A full connection layer at the end of convolutional networks is not necessary. Based on these studies, we propose a new convolutional architecture termed Three-Stream Convolutional Networks (TSCN). It first uses different streams to learn different aspects of feature maps for attentive spatio-temporal fusion of video, and then merges them together to study some union features. To further utilize the feature maps, two architectures are designed by using the strategies of multi-scale and upsampling. Comparative experiments on iLIDS-VID, PRID-2011 and MARS datasets illustrate that the proposed architectures are significantly better for feature extraction than the state-of-the-art models.

%（1）By utilizing filters with receptive fields of different sizes, the
%features learned by each column CNN are adaptive to variations in people/head size due to perspective effect or image
%resolution.
%----------这个是说多尺度的作用的，可以借鉴
%（2）应该有这条，虽然，单独使用平均池化或者步长为2并不好，但是，当我们联合使用的时候会得到很好的效果。
%（3）下采样在分类中起着重要的作用，因为数据如果一直很大的维数，就很难做分类等任务

\end{abstract}
%-------------------------------------------------------------------------

%%%%%%%%% BODY TEXT
\section{Introduction}

\begin{figure*}[t]
\begin{center}
%\fbox{\rule{0pt}{2in} \rule{0.8\linewidth}{0pt}}
   \includegraphics[width=0.8\linewidth]{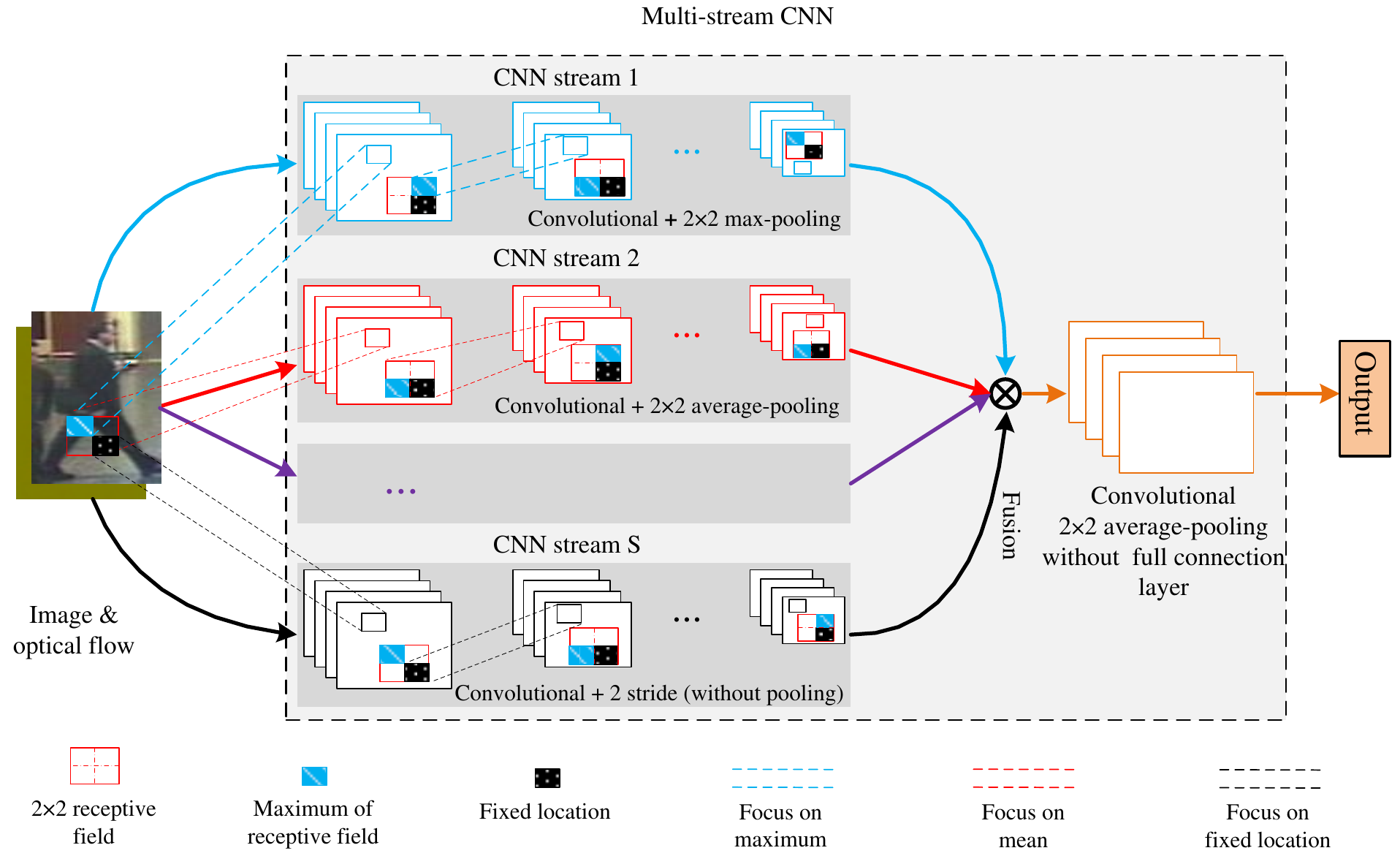}
\end{center}
   \caption{Example of multi-stream convolutional neural network. Different streams learn different aspects of the feature maps. Max-pooling, average-pooling and 2 stride convolutional layer focus on the maximum value, average value and fixed location of receptive field respectively.}
\label{fig:example}
\end{figure*}

The re-identification problem aims to identify the same person when he/she moves between non-overlapping cameras distributed at different locations \cite{gray2007evaluating}. It has received an increasing attention due to its potential applications such as people tracking in the surveillance videos and criminal investigation. However, it is still a very challenging task because person images/videos captured from the same/different cameras usually have large variations of lighting conditions, viewing points, body poses and backgrounds.

Currently, a variety of person re-identification algorithms have been developed. They can mainly be classified into two categories: still image-based approaches and the video-based ones. However, most of existing methods solve the person re-identification task with the former category \cite{cheng2016person,liu2017end,wang2016joint,wu2016personnet,xiao2016end,xiao2016learning}, while only a few methods are designed with the latter one. In reality, the video-based method is a more natural way to address the task of person re-identification. Intuitively, temporal information appeared in the videos can be used to capture the person motion. Moreover, videos contain rich samples of a person's appearance \cite{li2013learning,li2015delving}, which allow to build a better model with more discriminative poses, viewpoints, and backgrounds. Therefore, in this paper, we will focus on the problem of video-based person re-identification.

In a video-based method, optical flows are applied to extract temporal information from the consecutive frames of a
person. With the extracted optical flows, single-stream and two-stream architectures have been explored.
A single-stream architecture concatenates optical flows with RGB images as the inputs of a RNN (or CNN-RNN) model for training the model \cite{mclaughlin2016recurrent,xu2017jointly,zhang2017learning}. For a two-stream architecture, it separately builds two CNN architectures with optical flows and RGB images as their inputs, and then fuses/feeds them to a RNN model \cite{feichtenhofer2016convolutional,simonyan2014two,liu2017video}. As a result, all the single-stream and two-stream architectures can be viewed as methods of data fusion and model fusion, respectively. Instead of designing approaches for data fusion or model fusion, in this paper we will consider how to design effective architectures of networks that can fully utilize the learned feature maps.

Generally, the operation of pooling is used for downsampling in a CNN architecture. In practice, convolutional layers that have a stride of 2 also perform downsampling directly \cite{he2016deep}. It has been shown that max-pooling can achieve reasonable success on the task of video-based person re-identification \cite{mclaughlin2016recurrent,xu2017jointly,zhang2017learning}. However, when we use the max-pooling to sample the feature maps, it will dissipate many learned features. In fact, features that are not the maximum values also can help to solve the problem of person re-identification as shown in Section \ref{Sec:stream}.
%应该有这条，虽然，单独使用平均池化或者步长为2并不好，但是，当我们联合使用的时候会得到很好的效果。
In addition, the strategies of multi-scale and upsampling are also beneficial for reusing the learned features via mapping the features into different dimensionalities. Therefore, to make full use of the learned feature maps, we propose three new deep learning models that take the advantages of downsampling, multi-scale and upsampling. In each model, it contains multi-stream CNN architectures, where every stream focuses on different aspects of the learned feature maps.

%之后，CNN结构被输入到RNN或者RNN注意力机制模型中

Fig. \ref{fig:example} illustrates an example of multi-stream CNN architecture. As shown in this figure, we use different strategies to make the proposed model concentrate on different aspects of feature maps (e.g. max-pooling concentrates on the maximum value of receptive field). With these strategies, all the learned features will be fully used for feature representation on the task of video-based person re-identification. Because the best results are obtained by utilizing a three-stream CNN architecture, we refer it as three-stream convolutional network.

%Although each stream in networks with different streams has bad performance ,
%应该有这条，虽然，单独使用平均池化或者步长为2并不好，但是，当我们联合使用的时候会得到很好的效果。

%Although the stream such as average-pooling, convolutional layer with 2 stride gets bad performance, we can further improve the performance by using a mixture of them.

The rest of this paper is organized as follows. In Section \ref{Sec:related}, we will introduce the related work. Section \ref{Sec:Approach} will investigate the properties of different combinations, and then present the novel architectures. Section \ref{Sec:Results} will compare the performance of proposed method with other relevant state-of-the-art algorithms on iLIDS-VID, PRID-2011 and MARS datasets. Conclusions together with some further studies are summarized in the last section.

%第三段，现有视频方法都是基于光流和RGB图像的方法，他们也可以分为两类，一类是将光流法和RGB图像进行融合，即所谓的单流算法；
%另一类是将光流特征和RGB特征分别训练卷积 网络，然后，将两个学习到的网络进行融合。这是所谓的双流方法。在我们这篇论文中，我们研究的是单流算法，因为他将人的运动特征和
%图像特征合并进行处理，能整合一些分开学习所不能学习到的共同作用的特征。在此基础上，我们考虑了不同位置像素保留了不同的信息，
%现有的视频方法都是时空基于的方法根据处理数据方式的不同主要又可以分为两类：一种是，将光流和
%事实上，我们的三个流结构也可以运用到双流结构中。
%如图1所示，不同的流能保留不同的像素位置信息，因此，也能学到不同的特征。这就是三个流算法的成功之处。
%传统的最大池化或者平均池化，没能充分利用已有的特征信息，往往忽略了某些非最大化或者平均值的特征，因为单独考虑这些特征往往表现出
%不好的结果，因此，常常被忽略。
%在这里我们提出了三个流方法，如图1 所示。他能得到更多的特征信息，我们融合这些单独看起开不好的特征信息，提出了三个流的方法，之所以叫三个流算法，是因为，我们三个流的时候，发现效果是最好
%的，因此，我们就叫这个网络为三个流算法。进一步，我们将三个流的结构进行改进，提出了双流和多尺度三个流算法。

\section{Related Work}
\label{Sec:related}

In the past few years, researchers have designed various algorithms for person re-identification. These algorithms mainly focus on two aspects: feature representations \cite{ma2012local,liu2012person,zhao2014learning} and metric learning \cite{liao2015efficient,xiong2014person,zhang2015group,zhao2013person,paisitkriangkrai2015learning}. Gray and Tao \cite{gray2008viewpoint} learned viewpoint invariant features for pedestrian recognition by combining spatial and color information. Farenzena et al. \cite{farenzena2010person} extracted the texture histograms by studying the perceptual principles of symmetry and asymmetry. Kviatkovsky et al. \cite{kviatkovsky2013color} proposed a novel intradistribution structure based on the color distributions, which can learn the illumination invariant features under a variety of imaging conditions. Weinberger et al. \cite{weinberger2009distance} adopted the idea of large margin nearest neighbor metric (LMNN) to gather the $k$-nearest neighbors and separate examples from different classes by a large margin. Zheng et al. \cite{zheng2013reidentification} utilized relative distance  comparison (RDC) to maximize the likelihood of a pair of true matches. Liao et al. \cite{liao2015person} employed an effective architecture called Local Maximal Occurrence (LOMO) to learn a stable representation against viewpoint changes by maximizing the occurrence.

To use the temporal information, more and more researchers began to consider the video-based re-identification problem. With HOG3D descriptor, Klaser et al. \cite{klaser2008spatio} calculated the highest similarity between two video fragments as the distance of these two videos. Karaman et al. \cite{karaman2012identity} attempted to make the similar frames to have the similar labels by using a conditional random field (CRF).
Yan et al. \cite{yan2016person} tried to use the recurrent feature aggregation network (RFA-Net) to aggregate sequence level representations with LSTM.
More recently, McLaughlin et al. \cite{mclaughlin2016recurrent} applied a CNN to obtain the image-level representation and then fed it into a RNN to exploit the temporal information. Zhang et al. \cite{zhang2017learning} replaced the RNN with bidirectional recurrent neural networks (BRNN) to effectively learn spatio-temporal features. Zhou et al. \cite{zhou2017see} employed the spatial recurrent model (SRM) and temporal attention model (TAM) to study the spatio-temporal information. Xu et al. \cite{wu2016personnet} utilized spatial pyramid pooling and attentive temporal pooling to improve the performance. Liu et al. \cite{liu2017video} tried to accumulate the motion context by using a two-stream convolutional architecture. Unlike the existing deep learning based methods for video-based re-identification, our proposed models use a multi-stream convolutional architecture, in which different streams learn different aspects of feature maps and then they merge together to obtain some union characteristics.

%
%
%多列卷积网络也解释一下，说他是用于图像的，并且是只考虑到多尺度问题，没有考虑不同流关注不同问题。
%
%
%
%
%这段话很重要，可以仿照写一下：我们不采用一般的卷积网络，我们采用多流卷积网络
%In [2], features
%were extracted from each frame using a convolutional neural network that incorporated a recurrent final layer. Unlike
%[2][36], the temporal information is exploited in a bidirectional
%way in this paper. An end-to-end Bidirectional-RNN model
%is proposed and trained to exploit the temporal information
%before and after the current timestep and form a more complete
%representation about the sequential data
%
%
%
%
%
%最近几篇论文
%Recently, 双线性，CNN-RNN，ASPT，AOMC
%

%传统的静止图像基于的个人重识别方法;
%vedio基于的方法；
%重点讲一下深度学习的方法；
%然后，可以讲一下注意力机制模型。
%但是，讲一下，没有人考虑到如何利用所有特征的信息

\section{Approach}
\label{Sec:Approach}

A recurrent-convolutional network for video-based person re-identification usually consists of CNN architectures and a RNN network. The CNN architectures are utilized to extract feature representation from the multiple frames of video. The RNN network is applied to learn the temporal information between them. Downsampling is essential to a CNN architecture. As discussed previously, the traditional downsampling operation such as max-pooling only uses the maximum value, ignoring a lot of learned features. Therefore, in this paper, we propose the multi-stream convolutional network for video-based person re-identification. It can make full use of the learned feature maps by employing different streams of convolutional network which focus on different properties of the feature maps. To further reuse the learned features, two multi-scale convolutional networks are also developed.

\subsection{Different multi-streams}

%不同流的产生和作用

In this part, we consider two questions: 1) How to generate different multi-streams? 2) How to make different multi-streams focus on different aspects of the learned feature maps? The operation of max-pooling only focuses on the maximum value, wasting a lot of information. For example, if we use a 2$\times$2 max-pooling, only a quarter of the learned feature maps are adopted. Although the average-pooling considers all the feature maps, it treats all the features equally. Fortunately, a convolutional layer with a stride greater than 2 also performs downsampling by replacing the pooling operation. It concentrates on the learned feature maps with fixed positions. The dilated max/average-pooling also can be used for downsampling. With dilated max/average-pooling, the networks focus on larger receptive field. Recently, many pooling operations such as adaptive max/average-pooling and fractional max-pooling have been proposed. With these pooling operations, we can build multi-stream convolutional networks. Because different multi-streams focus on different aspects of the learned feature maps (e.g. max-pooling concentrates on the maximum value of receptive field), we can make full use of feature maps with these multi-streams. In fact, we can also generate different streams by using an ``upconvolutional'' layer, multi-scale or upsampling achieved by padding the smaller map with zeros. In the experiments, we will investigate most of the generating methods coupled with various of convolutional networks.

As shown in Section \ref{Sec:stream}, networks with different streams even including the one with the worst performance work better than networks with same streams, where each one has the best performance alone. It means that a stream with the poor performance can further improve the performance of the one with good result. In other words, different streams complement each other to learn some different aspects of the feature maps, that they can not learned independently. As a result, in our proposed model, three different streams will complement each other to learn some union features.

\subsection{Spatial fusion with multi-stream networks}
\label{Sec:fusion}

We consider different methods for fusing multi-stream convolutional networks. Because each network in the multi-streams concentrates on an aspect of the learned feature maps and it has the same channels and spatial resolution at the layers to be fused, we can simply stack layers on one network. To make it more concrete, we study several ways of fusing layers among multi-stream networks. We use $\mathbf{x}^{s} \in \Re^{H\times W \times D}$ to denote the feature maps of the $s$-th ($s=1,2,...,S$) stream, where $S$, $W$, $H$ and $D$ are the number of multi-streams, the width, height and channel number of the feature maps.  Let $f$ be the fusion function: $f(\mathbf{x}^{1},\mathbf{x}^{2},...,\mathbf{x}^{S})\longrightarrow \mathbf{y}$, where $\mathbf{y}$ is the fused feature map. $f$ can be easily used in any layer of the multi-streams if they have the same channels and spatial resolution.
\begin{itemize}
  \item \textbf{Sum fusion.} In a multi-stream network, the sum fusion, $\mathbf{y}^{sum} =f^{sum}(\mathbf{x}^{1},\mathbf{x}^{2},...,\mathbf{x}^{S_{1}})$, is to compute the sum of some or all the feature maps at the same spatial locations $i,j$ and channels $d$:
\begin{equation}
\begin{aligned}
y_{i,j,d}^{sum} = x_{i,j,d}^{1}+x_{i,j,d}^{2}+,...,+x_{i,j,d}^{S_{1}},
\end{aligned}
\end{equation}
where $1\leq i \leq H, 1\leq j \leq W, 1\leq d \leq D, 2\leq S_{1}\leq S$ and $\mathbf{x}^{1},\mathbf{x}^{2},...,\mathbf{x}^{S_{1}},  \mathbf{y}^{sum} \in \Re^{H\times W \times D}$.

  \item \textbf{Max fusion.} Similarly, the max fusion, $\mathbf{y}^{max}  =f^{max}(\mathbf{x}^{1},\mathbf{x}^{2},...,\mathbf{x}^{S_{1}})$, takes the maximum of some or all the feature maps:
\begin{equation}
\begin{aligned}
y_{i,j,d}^{max} = \rm{max}  \{x_{i,j,d}^{1},x_{i,j,d}^{2},...,x_{i,j,d}^{S_{1}}\},
\end{aligned}
\end{equation}

  \item \textbf{Channel fusion.} In a multi-stream network, the channel fusion, $\mathbf{y}^{cha} =f^{cha}(\mathbf{x}^{1},\mathbf{x}^{2},...,\mathbf{x}^{S_{1}})$, stacks some or all the feature maps at the same spatial locations $i,j$ across channels $d$:
\begin{equation}
y_{i,j,t}^{cha} =x_{i,j,d}^{s},
\end{equation}
\begin{equation}
s.t \quad \left\{
             \begin{array}{ll}
             s=1, &   1 \leq t \leq D \\   \nonumber
             s=2, &   D+1 \leq t \leq 2D \\
             ... &  \\
             s=S_{1}, &   (S_{1}-1) \times D+1 \leq t \leq S_{1} \times D
             \end{array}
\right.
\end{equation}
where $\mathbf{y}^{cha} \in \Re^{H\times W \times (S_{1} \times D)}$.

  \item \textbf{Width fusion.} In a multi-stream network, the width fusion, $\mathbf{y}^{wid} =f^{wid}(\mathbf{x}^{1},\mathbf{x}^{2},...,\mathbf{x}^{S_{1}})$, stacks some or all the feature maps at the same spatial heights $i$ and channels $d$ across spatial widths $j$:
\begin{equation}
y_{i,t,d}^{wid} =x_{i,j,d}^{s},
\end{equation}
\begin{equation}
s.t \quad \left\{
             \begin{array}{ll}
             s=1, &   1 \leq t \leq W \\   \nonumber
             s=2, &   W+1 \leq t \leq 2W \\
             ... &  \\
             s=S_{1}, &   (S_{1}-1) \times W+1 \leq t \leq S_{1} \times W
             \end{array}
\right.
\end{equation}
where $\mathbf{y}^{wid} \in \Re^{H\times (S_{1} \times W) \times D}$.

   \item \textbf{Height fusion.} In a multi-stream network, the height fusion, $\mathbf{y}^{hig} =f^{hig}(\mathbf{x}^{1},\mathbf{x}^{2},...,\mathbf{x}^{S_{1}})$, stacks some or all the feature maps at the same spatial widths $j$ and channels $d$ across spatial heights $i$:
\begin{equation}
y_{t,j,d}^{hig} =x_{i,j,d}^{s},
\end{equation}
\begin{equation}
s.t \quad \left\{
             \begin{array}{ll}
             s=1, &   1 \leq t \leq H \\   \nonumber
             s=2, &   H+1 \leq t \leq 2H \\
             ... &  \\
             s=S_{1}, &   (S_{1}-1) \times H+1 \leq t \leq S_{1} \times H
             \end{array}
\right.
\end{equation}
where $\mathbf{y}^{hig} \in \Re^{(S_{1} \times H) \times  W \times D}$.
\end{itemize}

\begin{figure}[t]
\begin{center}
%\fbox{\rule{0pt}{2in} \rule{0.8\linewidth}{0pt}}
   \includegraphics[width=1.0\linewidth]{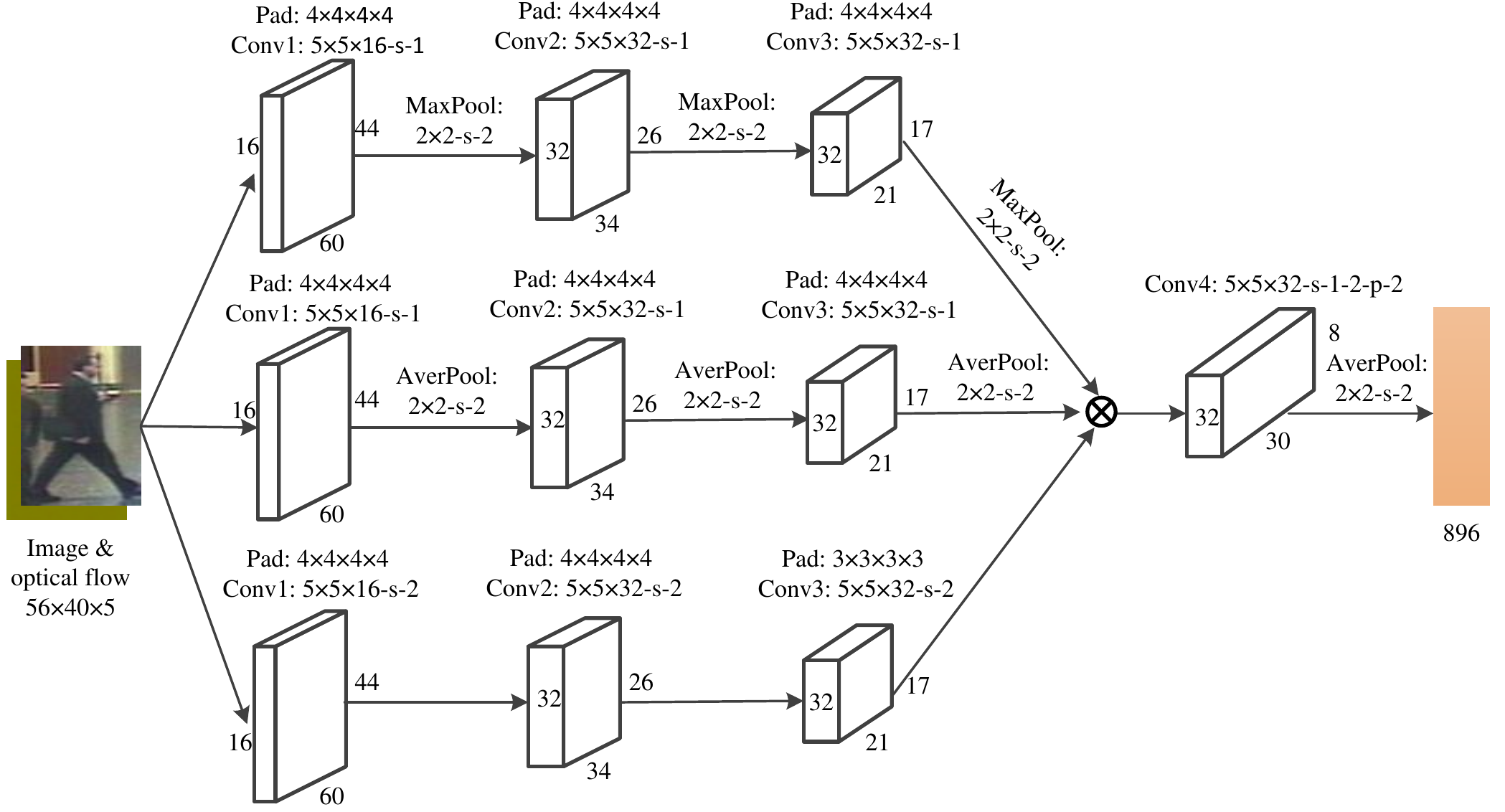}
\end{center}
   \caption{The architecture of three-stream convolutional network. The inputs consist of RGB images and optical flows. We use it as the baseline model for choosing parameters.}
\label{fig:base}
\end{figure}

With the baseline model as shown in Fig. \ref{fig:base}, we evaluate and compare each possible fusion method in terms of their classification accuracy in our experiments. Note that the number of channel is arbitrary for channel fusion. We can change the number of channel on any stream and optimize over the filters at any layer to make this arbitrary correspondence useful for subsequent learning. In the case of width fusion, the width number of the feature maps can also be changed. Similarly to width fusion, the height number of the feature maps is again arbitrary for height fusion. For convenience, we fix the width, height and channel number of the feature maps for all the layers to be fused in the experiments.

Spatial fusion can be applied at any point among the multi-stream networks when they have same spatial resolution. Actually, the spatial fusion have significant impact on the number of parameters and layers. In the experimental section, we evaluate some networks that spatial fusion can be placed at different points to implement e.g. early-fusion or late-fusion.

In this paper, we show that it is not necessary to use a full connection layer in the multi-stream convolutional networks, even it's harmful in a recurrent-convolutional network. Many previous work have been shown that a full connection layer with small size of 128 is necessary at the end of convolutional networks and it can achieve the state-of-the-art performance. However, in the multi-stream networks, we replace the full connection layer with a convolutional layer.  As a result, the image-level representation is more meaningful and the dimension of feature space can be set to a very high value reserved more information of the image. Without a full connection layer at the end of convolutional networks, we obtain the state-of-the-art performance.

\subsection{Multi-stream with multi-scale and upsampling}

As mentioned above, we can easily implement spatial fusion at any point among the multi-stream networks, with the only constraint that the feature maps have the same spatial resolution. This can be achieved by using multi-scale, upsampling or an ``upconvolutional'' layer. Multi-scale can be applied to mapping the features into different dimensionalities. Upsampling can simply be implemented by padding the smaller feature map with zeros. However, we do not utilize the ``upconvolutional'' layer in our models. In practice, multi-scale and upsampling are adaptive to variations in person/body size due to perspective effect or image resolution in the frames of the video.

\begin{figure}[t]
\begin{center}
%\fbox{\rule{0pt}{2in} \rule{0.8\linewidth}{0pt}}
   \includegraphics[width=1.0\linewidth]{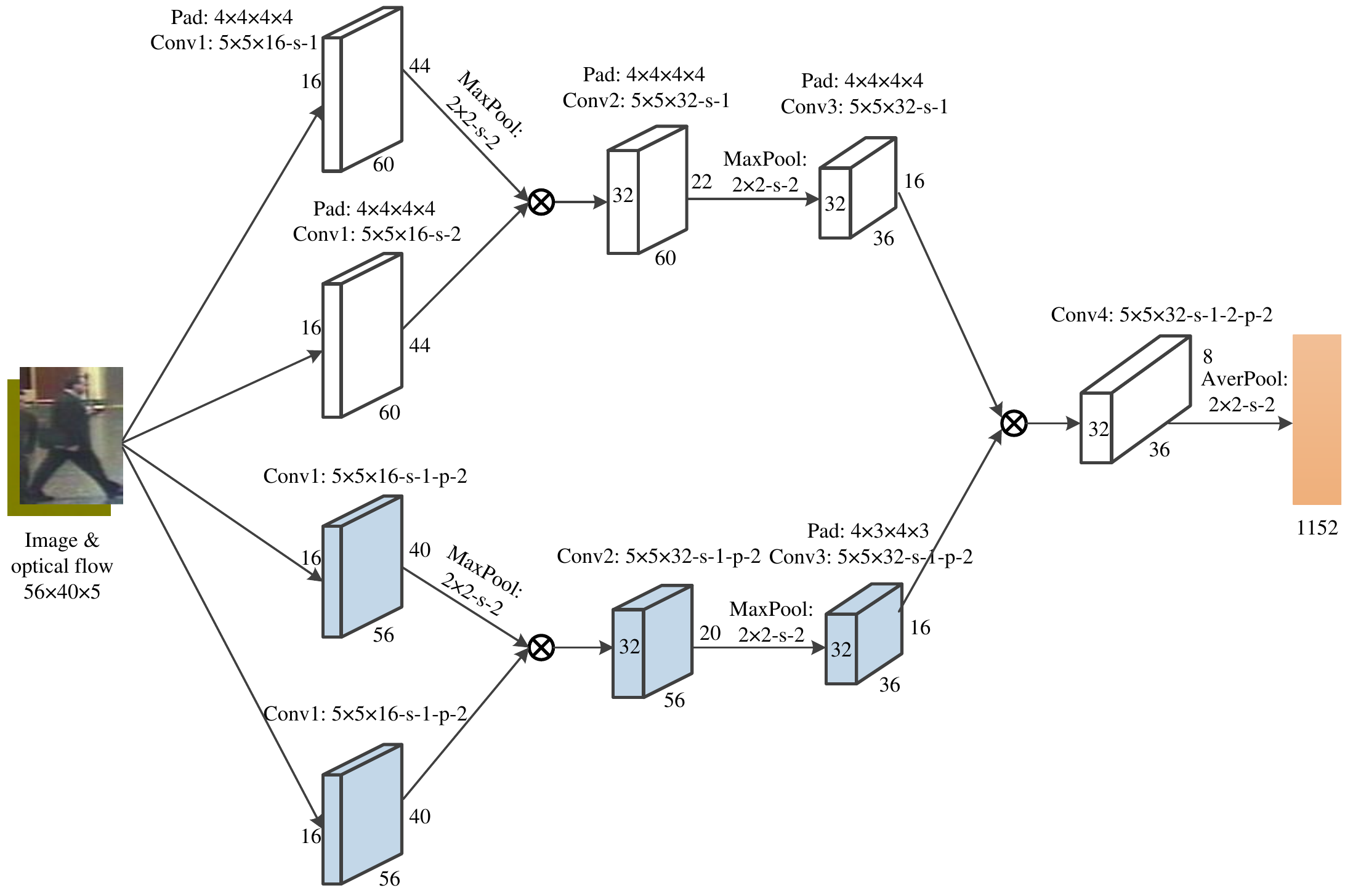}
\end{center}
   \caption{The architecture of two-stream multi-scale convolutional network. It has 3 width fusion layers.}
\label{fig:stream2}
\end{figure}

\begin{figure}[t]
\begin{center}
%\fbox{\rule{0pt}{2in} \rule{0.8\linewidth}{0pt}}
   \includegraphics[width=1.0\linewidth]{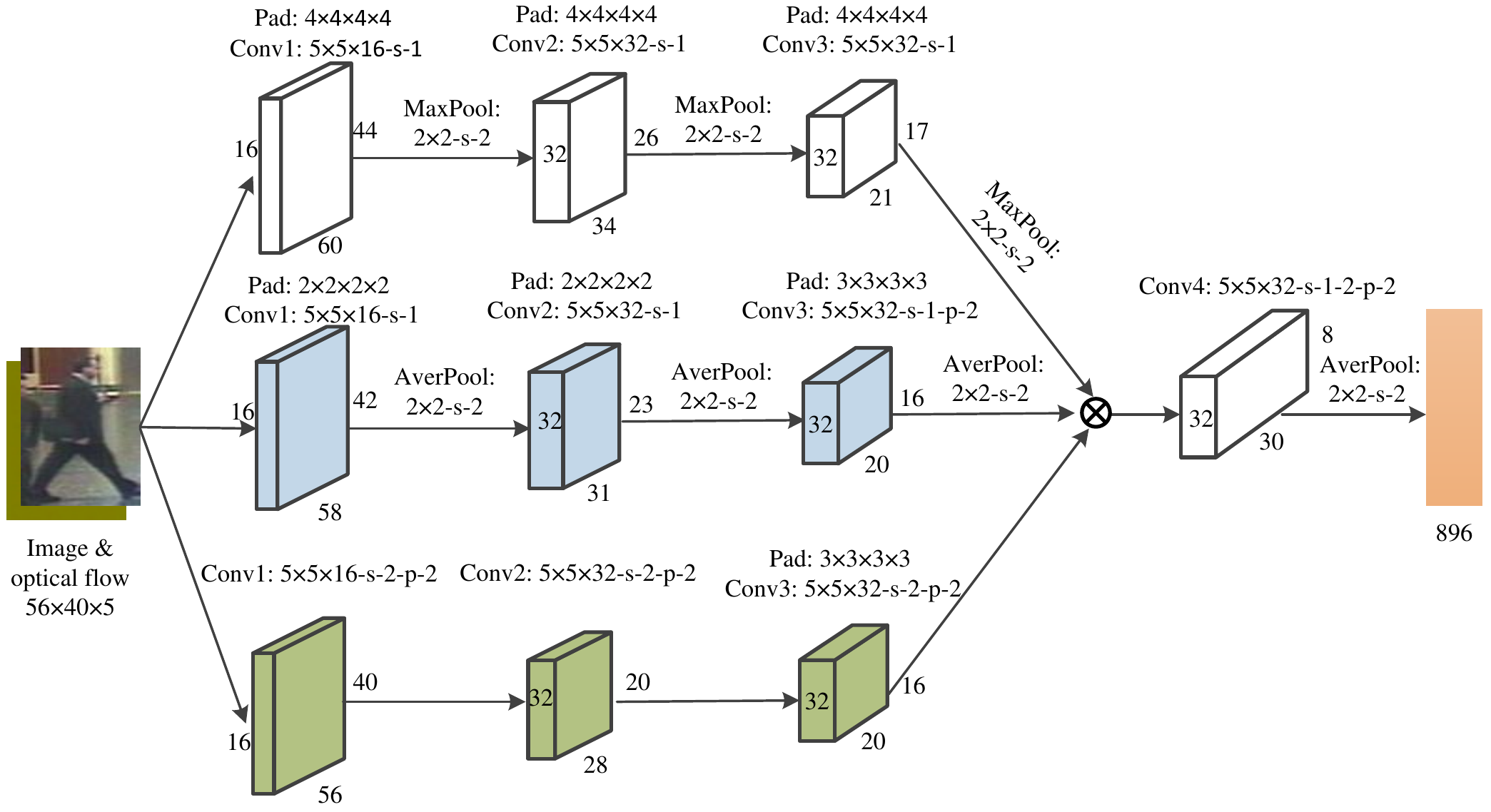}
\end{center}
   \caption{The architecture of three-stream multi-scale convolutional network.}
\label{fig:multi3}
\end{figure}

With multi-scale and upsampling, we construct two multi-stream CNN architectures: two-stream multi-scale network and three-stream multi-scale network. Similarly to pooling operations, networks with multi-scale focus on some different characteristics of the feature maps. As shown in Fig. \ref{fig:stream2}, the two-stream multi-scale network can be fused at three layers, which can achieve the goal of pixel-wise registration of the channels from each stream. The three-stream multi-scale network uses multi-scale to obtain different dimensionalities at the lower-level layers and gets the same spatial resolution with upsampling before the fusion (higher-level) layer, see Fig. \ref{fig:multi3}.

%\subsection{Attentive temporal pooling}
%%这里要讲几个时间融合的方法，然后，我们选择了一个好的――这里可以不用单独列出，不是我们的重点
%%需要写，简单写一段话，说我们调查了不同的时间融合方式，并给出了对比实验
%
%Similar to the method
%The recurrent layer is able to capture temporal information with hidden states.
%我们的工作在用充分利用已学习到的特征，这部分工作不是我们的。但为了流程的完整性，我们采用了目前最好的时间池化方法，具体可以
%参考论文。他采用
%RNN这个有重要作用，但是运用在这里有两个不足，幸好等人解决了这样的不足。但是原始的这样信息却包含了很多无用信息，
%注意力池化可以解决，因此，我们采用最这样的方法在我们的时间池化中。

\subsection{Proposed architecture}
%第一段Based on the 先前的讨论，
%我们根据前面的内容提出了一个新的模型，这个模型的参数最优的选择是基于实验的。
%第二段
%这个结构的总体情况如图3所示。他由两块组成：CNN+RNN。然后，说一下整个流程，用语言描述每个模块，以及整个过程

Based on the previous discussion, we propose a new attentive spatio-temporal fusion architecture. In fact, it can be extended to three new attentive spatio-temporal fusion models by replacing the part of multi-stream CNN with three effective CNN architectures shown in Figs. 2-4. The choices of the parameters for proposed architectures (e.g. number of stream, spatial fusion method, layer and attentive temporal pooling) are based on our empirical evaluation.

\begin{figure}[t]
\begin{center}
%\fbox{\rule{0pt}{2in} \rule{0.8\linewidth}{0pt}}
   \includegraphics[width=1.0\linewidth]{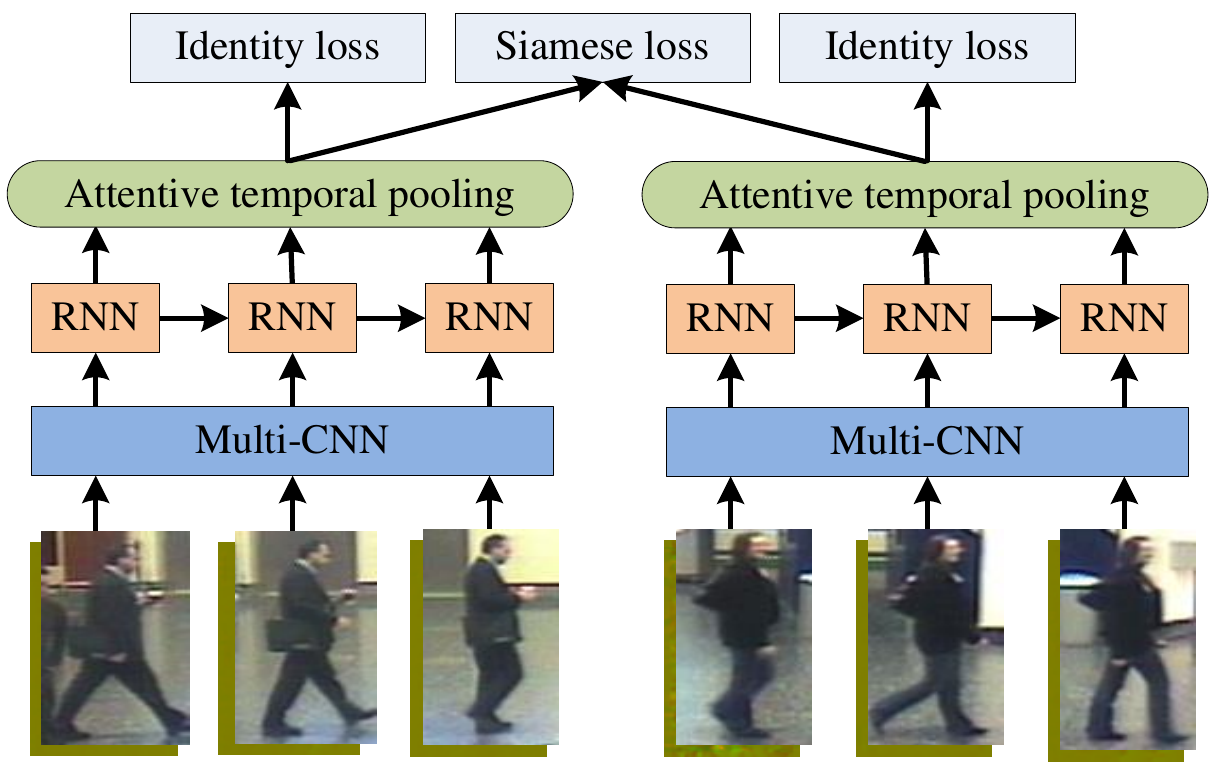}
\end{center}
   \caption{The proposed architecture for video-based person re-identification.}
\label{fig:art}
\end{figure}

Fig. \ref{fig:art} illustrates the architecture of the proposed network. It employs a Siamese network architecture, which has two sub-networks with same weights. As shown in this figure, our network architecture consists of three parts: convolutional network, the recurrent network and the attentive temporal pooling layer. The recurrent network is used for capturing temporal information of the frames in a video and the attentive temporal pooling layer inspired by the works of \cite{xu2017jointly} guides for effectively extracting this temporal information. For convolutional network, it is a critical part of the attentive spatio-temporal fusion architectures. We design three multi-stream convolutional networks to replace the single-stream or two-stream architectures. As mentioned above, the multi-stream convolutional networks can make full use of the feature maps because each stream focuses on some characteristics of the feature maps.

As suggested by  \cite{mclaughlin2016recurrent}, both the Siamese loss and the identity loss are used to train the proposed architectures. Given a pair of sequences of persons $i$ and $j$, we use the Siamese network to obtain the sequence feature vectors $v_{i}$ and $v_{j}$. After that, the Siamese loss objective function with Euclidean distance can be given as follows:
\begin{equation}
E(v_{i},v_{j})=\left\{
             \begin{array}{ll}
             ||v_{i}-v_{j}||_{2}^{2}, &    i=j\\
             \rm{max}\{0,m-||v_{i}-v_{j}||_{2}^{2}\}, &    i\neq j \\
             \end{array}
\right.
\end{equation}
where $m$ is the margin that separates features of different persons. In addition, we use standard softmax function to predict the identity of the person in the sequence, and then we adopt the cross-entropy loss to obtain the identity loss objective function $L(v_{i})$ and $L(v_{j})$. Finally, we define the overall training objective function $P(v_{i},v_{j})$ by simultaneously optimizing the Siamese loss and the identity loss:
\begin{equation}
P(v_{i},v_{j})=E(v_{i},v_{j})+L(v_{i})+L(v_{j})
\end{equation}

Here, we treat the Siamese loss and the identity loss equally. The proposed architectures can be trained end-to-end using back-propagation-through-time. We detail the training parameters in the next section.

%\subsection{Model details}
%
%1) Input
%As the human body only covers a small part of a surveillance video and normally has a small image size of 128 × 64
%in most current datasets, it is unnecessary to use some largesized CNNs like VGG and GoogleNet which may result in
%overfitting. Also, considering that the amount of parameters in
%BRNN is not large, the adopted CNN should be comparable to
%maintain the balance between them. Similar to McLaughlin’s
%work [2], the network with three convolutional layers and three
%max-pooling layers is adopted. When performing convolution
%on the image or the feature map, we first pad the image
%or feature map with four numbers of zeros. The hyperbolictangent (Tanh) is used as non-linear activation-function after
%each convolution layer. The architecture of CNN is shown in
%Figure 2
%
%In detail, we first resize the pedestrian frames from 128×64
%to 64 × 48 [2], then randomly crop patches of size 56 × 40
%from the rescaled images and feed them into the CNN.
%Finally, after three-layers convolution and pooling, a 2560
%dimensional feature vector is produced as the representation
%of a single frame.
%
%3) Data Augmentation

\section{Experimental Results}
\label{Sec:Results}

In this section, we evaluate our proposed models for video-based person re-identification on iLIDS-VID  \cite{wang2014person}, PRID-2011 \cite{hirzer2011person} and MARS \cite{zheng2016mars} datasets and compare the performance with state-of-the-art algorithms. Several important parameters will also be experimentally evaluated.

\subsection{Datasets}

The iLIDS-VID dataset contains 300 persons, where each person is represented by two sequences appeared in two non-overlapping camera views at an airport arrival hall under a multi-camera CCTV network. The length of sequences range from 23 to 192 frames with an average length of 73. Due to clothing similarities for different persons, lighting and viewpoint variations, cluttered background and random occlusions, it becomes very challenging.

The PRID-2011 dataset consists of 749 persons captured by two non-overlapping cameras. Each image sequence has the length of frame from 5 to 675, with an average number of 100. Compared with the iLIDS-VID dataset, it has simple backgrounds and rare occlusions. Following the protocol used in \cite{xu2017jointly}, only the first 200 persons captured by both cameras are utilized.

The MARS dataset is the largest video-based person re-identification benchmark dataset to date. It has 1261 different persons, each person has at least two image sequences automatically obtained by DPM detector and GMMCP tracker. These sequences are captured by 2-6 cameras and each identity has 13.2 sequences on average. Similar to the protocol used in \cite{xu2017jointly}, we randomly choose two cameras of the same person for evaluation, where the case was reduced to experiences with iLIDS-VID and PRID-2011.

%\subsection{Experiment settings}

Following \cite{xu2017jointly}, we randomly split each dataset into training set and testing set with equal size. We repeat the experiments 10 times with  different splits and report the average results with Cumulative Matching Characteristics (CMC) curves. To train the Siamese network, we set the margin to 4 and choose the sequences of the same person or different person under different cameras as the positive or negative pairs respectively. For the fairness of experiments, we set the length of each person sequence to 16 for training and 128 for testing. We set the initial learning rate to 2e-3, and multiply it by 0.5 after the 800th epoch, and accomplish the training process at 1200th epoch.  Given 150 persons, training for 1200 epochs takes about one day and a half using the Nvidia K80 GPU.

\subsection{Performance evaluation of multi-streams}

Before comparing the performance with the state-of-the-art methods, we conduct several experiments on iLIDS-VID dataset to verify the effectiveness of our proposed multi-stream networks.

\begin{table}
  \centering
    \caption{Performance comparison of two-stream networks.}
\begin{tabular}{|c|c|c|c|c|c|}
  \hline
Dataset & \multicolumn{4}{|c|}{iLIDS-VID}\\
     \hline
        Streams         & R=1 & R=5  &   R=10 &   R=20\\
        \hline \hline
        MaxPool + MaxPool                & 65.1    & 88.1 &  95.6  & 98.4  \\
        \hline AverPool + AverPool         & 63.2    & 87.2 &  95.1  &97.5 \\
        \hline 2 stride + 2 stride       & 62.5    & 85.4 & 93.6   & 96.3                     \\
        \hline DilatedMax + DilatedMax   & 64.5    & 87.2 & 95.3   & 97.8                    \\
        \hline  MaxPool + AverPool        & 65.6    & 88.9 & 95.4   & 98.6                  \\
        \hline  MaxPool + 2 stride       & 65.3    & 88.2 & 95.8   & 98.6                    \\
        \hline  MaxPool + DilatedMax     & 65.4    & 88.1 & 95.4   & 98.7                     \\
        \hline  AverPool + 2 stride       & 64.1    & 87.6 & 95.2   & 97.7                      \\
        \hline  AverPool + DilatedMax     & 64.8    & 88.1 & 95.6   & 98.1                      \\
        \hline  2 stride + DilatedMax    & 64.6    & 87.6 & 95.1   & 97.6                        \\
    \hline
\end{tabular}
\label{tab:mix}
\end{table}

\subsubsection{Networks with different streams vs. networks with same streams}
\label{Sec:stream}

To show the effectiveness of multi-stream networks, we construct two types of multi-stream architectures: networks with different streams and networks with same streams. As mentioned above, we can generate a different stream by utilizing any method such as max-pooling, average-pooling, convolutional layer with 2 stride, dilated max/average-pooling, adaptive max/average-pooling, fractional max-pooling, ``upconvolutional'' layer, multi-scale and upsampling. For simplicity, here we use two-stream networks to test the effectiveness of multi-stream networks. Therefore, we can get the networks with different streams by using any two generating ways. To construct the networks with same streams, we use any one of the generating ways to obtain two same sub-networks. For comparing, one of the stream for these two multi-stream architectures is identical.

The average accuracy of the comparison for these two types of multi-stream architectures is reported in Table \ref{tab:mix}. Because of large results of the combination for any two generating ways, we only test some common generating ways. It is sufficient to illustrate this phenomenon: networks with different streams work better than networks with same streams among the testing architectures. As shown in the table, the architecture with two same max-pooling sub-networks gets the best result among the testing networks with same streams. The architecture constructed with two same 2 stride sub-networks obtains the worst performance. However, when we use both max-pooling and convolutional layer with 2 stride to construct a network with different streams, we can obtain very good results. It means that a stream with the poor performance can further improve the performance of the one with good result. In other words, different streams complement each other to learn some different aspects of the feature maps, that they can not learn independently.

%
%\begin{table}
%  \centering
%  \begin{tabular}{|c|p{3cm}|cc|}
%        \hline a & bbb&22&88\\
%        \hline a & This is a very long sentence that may exceed the bound of this table.&22&88 \\
%        \hline
%  \end{tabular}
%\end{table}

%1) Why multi-stream with different stream works better？ （表1）
%实验说明：最大池化，单独使用可能是最好的，但合并起来，混合使用才是更好的。这说明不，
%------不同流能相补充，使得效果更好
%另一方面，说明表现（performance）不好的流，
%在加入到好的流中，也能帮助 （进一步） 提高准确性――总之，这就说明了，不同流能学到不同的侧面，（进而），the学习到的侧面有助于提高性能。

\subsubsection{Which is the best number of streams?}

\begin{figure}[t]
\begin{center}
%\fbox{\rule{0pt}{2in} \rule{0.8\linewidth}{0pt}}
   \includegraphics[width=0.8\linewidth]{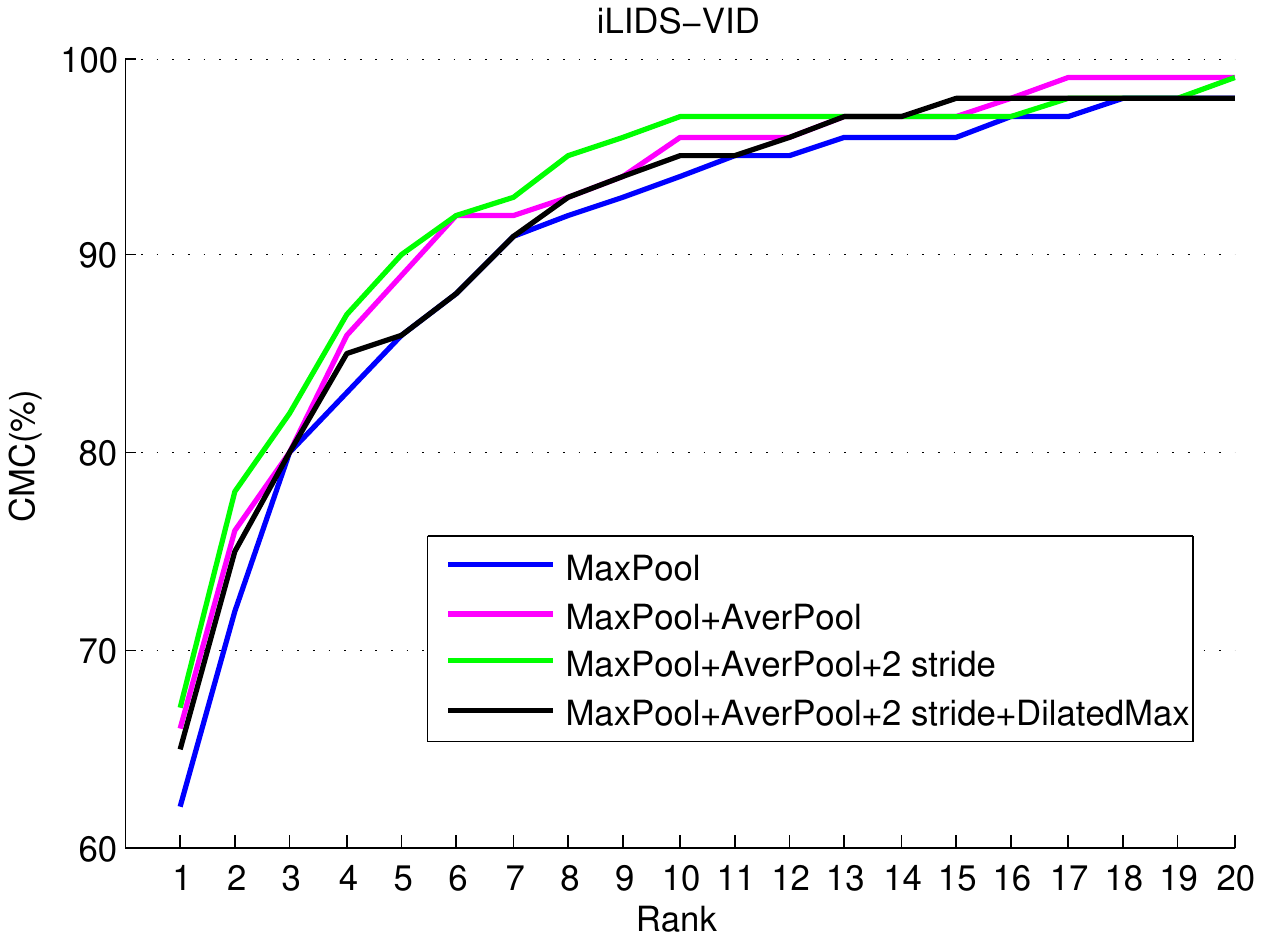}
\end{center}
   \caption{Performance comparison with different number of stream.}
\label{fig:layernumber}
\end{figure}

We construct four networks with increasing the number of streams from 1 to 4. We use the network of \cite{xu2017jointly} as the one-stream architecture. The two-stream architecture is composed of max-pooling and average-pooling. We add the convolutional layer with 2 stride into two-stream network to form the three-stream architecture and the four-stream architecture consists of max-pooling, average-pooling, convolutional layer with 2 stride and dilated max-pooling. As shown in Fig. \ref{fig:layernumber}, the three-stream architecture gets the best performance, this is why we refer our proposed multi-stream networks as three-stream convolutional network.

\subsubsection{How and where to fuse multi-stream networks?} %（表3）

We investigate how and where to fuse multi-stream networks. The three-stream baseline architecture is used as the testing model in this part (See Fig. \ref{fig:base} for details). As the fusion layer can be injected at any location and two or more fusion layers can be adopted in a three-stream network, the combination of them has many options. For example, with two fusion layers,  we can first fuse two streams at a certain layer and then merge with the third one at another layer. The combination of two streams has 3 situations and the first fusion layer can be located at any layer. As a result, for a 3 layers network, it has at least 6 combinations. Hence, we only consider that one fusion layer is used in the three-stream network.
\begin{table}
  \centering
    \caption{Performance comparison for different fusion methods. The results are tested on the baseline model as shown in Fig. \ref{fig:base}.}
\begin{tabular}{|c|c|c|c|c|c|}
  \hline
Dataset & \multicolumn{4}{|c|}{iLIDS-VID}\\
     \hline
        Fusion methods         & R=1 & R=5  &   R=10 &   R=20\\
        \hline \hline
         Sum       & 62.6   & 85.2 & 94.7  & 96.8    \\
        \hline Max       & 64.7   & 86.6 & 94.8  & 97.3         \\
        \hline Channel   & 65.4   & 87.8 & 95.3   & 97.4 \\
        \hline Width     & 66.5  & 89.5 & 96.6 & 98.2         \\
        \hline  Height    & 66.1   & 88.9 & 96.2   & 97.8      \\
    \hline
\end{tabular}
\label{tab:fusion}
\end{table}

To choose the best method of fusing layers among multi-stream networks, we compare different fusion strategies with the three-stream baseline architecture. Table \ref{tab:fusion} reports the performance for all the fusion methods as described in Section \ref{Sec:fusion}. We observe that width fusion performs the best and is slightly better than height fusion. We also see that the sum fusion gets the worst result. It is not surprising as the sum fusion adds three streams together. As width fusion obtains the best performance, we use it for testing and compare the performance for fusion from different layers in Table \ref{tab:where}. As shown in this table, fusing the three-stream network at Conv3 achieves the best performance.

\begin{table}
  \centering
    \caption{Performance comparison for Conv fusion at different fusion layers. The results are also tested on the baseline model with width fusion.}
\begin{tabular}{|c|c|c|c|c|c|}
  \hline
Dataset & \multicolumn{4}{|c|}{iLIDS-VID}\\
     \hline
        Fusion layers         & R=1 & R=5  &   R=10 &   R=20\\
        \hline \hline
        Conv1       & 62.3   & 86.4 & 94.8  & 96.7    \\
       \hline Conv2       & 65.7   & 88.5 & 96.1  & 97.8         \\
       \hline Conv3    & 66.5  & 89.5 & 96.6 & 98.2    \\
       \hline Conv4   & 61.4   & 84.3 & 93.8   & 96.1 \\
    \hline
\end{tabular}
\label{tab:where}
\end{table}

\subsection{Comparison with State-of-the-Art Methods}

To further evaluate the performance of multi-stream networks, we compare the proposed three architectures with the state-of-the-art methods on iLIDS-VID, PRID-2011 and MARS datasets.

\begin{table*}
  \centering
    \caption{Comparison of our approaches with other state-of-the-art methods on iLIDS-VID and PRID-2011. Note that the better results of AMOC are obtained by using a better optical flow algorithm \cite{revaud2015epicflow}. With the old algorithm \cite{lucas1981iterative}, our proposed models achieve the best performance against all the methods including AMOC.}
\begin{tabular}{|c|c|c|c|c|c|c|c|c|c|c|c|}
  \hline
\multicolumn{2}{|c|}{Dataset}   & \multicolumn{4}{|c|}{iLIDS-VID} & \multicolumn{4}{|c|}{PRID-2011}\\
     \hline
        Methods & Years   & R=1 & R=5  &   R=10 &   R=20  & R=1 & R=5  &   R=10 &   R=20\\
        \hline \hline
        STA \cite{liu2015spatio}              &  2015      & 44.3 & 71.7 & 83.7 & 91.7       & 64.1 & 87.3 & 89.9 & 92.0  \\
       \hline DVR \cite{wang2014person}        &2014      &39.5 & 61.1 & 71.7 &81.8          &40.0 & 71.7 & 84.5 & 92.2  \\
       \hline SRID \cite{karanam2015sparse}    &  2015     &24.9 & 44.5 & 55.6 & 66.2        &35.1 & 59.4 & 69.8 &79.7    \\
       \hline AFDA \cite{li2015multi}           & 2015      &37.5 & 62.7 & 73.0 & 81.8        &43.0 & 72.7 & 84.6 &91.9 \\
      \hline   DVDL \cite{karanam2015person}      & 2015      & 25.9 & 48.2 & 57.3 & 68.9      & 40.6 &69.7 & 77.8 &85.6   \\
       \hline CNN-RNN \cite{mclaughlin2016recurrent} & 2016           & 58.0 & 84.0 & 91.0 & 96.0      & 70.0 & 90.0 & 95.0 & 97.0           \\
        \hline  CNN-BRNN \cite{zhang2017learning}  &2017       &55.3 & 85.0 & 91.7 & 95.1         & 72.8 & 92.0 & 95.1 &97.6\\
       \hline ASTPN \cite{xu2017jointly}       & 2017          &62.0 & 86.0 & 94.0 & 98.0       & 77.0 & 95.0 & 99.0 & 99.0     \\
       \hline AMOC + EpicFlow \cite{liu2017video}      & 2017    & 68.7 & 94.3 & 98.3 & 99.3       & 83.7 & 98.3 & 99.4 & 100    \\
       \hline AMOC + LK-Flow  \cite{liu2017video}       & 2017  & 65.3 & 87.3 & 96.1 & 98.4       & 78.0 & 97.2 & 99.1 & 99.7 \\
       \hline our TSCN              &-          & 66.5  & 89.5 & 96.6 & 98.2      & 79.2  & 97.4 & 99.5 & 100  \\
       \hline our Multi-TSCN        &-          & 67.5  & 90.4 & 97.2 & 98.6      & 78.8  & 96.7 & 99.1   & 99.6  \\
       \hline Multi-Two-SCN        &-          & 65.4  & 87.8 & 96.2 & 97.5      & 79.7  & 97.5  &  99.2 & 99.9 \\
    \hline
\end{tabular}
\label{tab:state}
\end{table*}

\subsubsection{Results on iLIDS-VID and PRID-2011}

We now compare the performance of our proposed models with the state-of-the-art methods for video-based re-identification: STA \cite{liu2015spatio}, DVR \cite{wang2014person}, SRID \cite{karanam2015sparse}, AFDA \cite{li2015multi}, DVDL \cite{karanam2015person}, CNN-RNN \cite{mclaughlin2016recurrent}, CNN-BRNN, ASTPN \cite{zhang2017learning} and AMOC \cite{liu2017video}. Note that the better results of AMOC are obtained by using a better optical flow algorithm \cite{revaud2015epicflow}. With the old algorithm \cite{lucas1981iterative}, our proposed models achieve the best performance against all the methods including AMOC on both iLIDS-VID and PRID-2011 datasets.

The CMC results of our architectures with other models are listed in Table \ref{tab:state}. In general, networks with different multi-streams perform better than other networks without this architecture on both iLIDS-VID and PRID-2011 datasets. With old algorithm of optical flow, our proposed two-stream multi-scale model can achieve a matching rate of rank-1 of about 65.4\% on iLIDS-VID dataset, which is higher than all the testing methods. When we use three-stream multi-scale architecture, the performance can be further improved, especially for rank-1 and rank-5. The improvements are 2.1\% and 2.6\% for rank-1 and rank-5 respectively. For PRID-2011 dataset, the two-stream multi-scale network outperforms other methods, with rank-1 accuracy of 79.7\%. It is possible that the two-stream multi-scale network has less parameters and can be trained well on the less challenging dataset. The reason that our proposed models achieve such good performance is that different multi-streams focus on different aspects of the feature maps and they complement each other to learn some union characteristics.

%Comparing the CMC results shown in Tab.
%
%
%
%
%
%Their great improvement is achieved by using a better (but offline) optical flow algorithm
%and a two-stream
%architecture where motion and appearance features are better
%modelled by learning them separately, fusing them before
%
%
\begin{table}
  \centering
    \caption{The CMC Rank accuracy on MARS (\%).}
\begin{tabular}{|c|c|c|c|c|c|}
  \hline
Dataset & \multicolumn{4}{|c|}{MARS}\\
     \hline
        Methods         & R=1 & R=5  &   R=10 &   R=20\\
        \hline \hline
         RNN-CNN  \cite{mclaughlin2016recurrent}      & 40.0  & 64.0  & 70.0  & 77.0   \\
        \hline ASTPN \cite{xu2017jointly}      & 44.0  & 70.0 &   74.0  & 81.0  \\
        \hline Ours        & 45.6  & 72.4 &   75.4  & 82.6   \\
    \hline
\end{tabular}
\label{tab:mars}
\end{table}

\subsubsection{Results on MARS}

To further evaluate the proposed architectures, we also conduct experiments on the large and realistic MARS dataset. We use the similar protocol in \cite{xu2017jointly} and randomly choose two cameras of the same person for testing. Table \ref{tab:mars} presents the performances of our model compared with the RNN-CNN and ASTPN \cite{xu2017jointly}. As shown in Table \ref{tab:mars}, our proposed model still achieves the best accuracy. It again illustrates the effectiveness of our proposed architecture that uses different streams to take full use of feature maps. It should be pointed out that we only test the two-stream multi-scale architecture with max-pooling and dilated max-pooling for this dataset.

%Their great improvement is achieved by using a better (but offline) optical flow algorithm
%and a two-stream
%architecture where motion and appearance features are better
%modelled by learning them separately, fusing them before
%the RNN layer.

%\subsection{Cross-dataset testing}

%\begin{table}
%  \centering
%    \caption{Results.   Ours is better.}
%\begin{tabular}{|c|c|c|c|c|c|c|}
%     \hline
%        Model  & Trained on       & R=1 & R=5  &   R=10 &   R=20\\
%        \hline \hline
%        Baseline       & iLIDS-VID  & 32.5 & 60.4   & 73.2     & 86.7       \\
%       \hline Two-         & iLIDS-VID    & 32.9 & 61.2   & 73.6     & 86.8         \\
%       \hline Three       & iLIDS-VID     & 32.1 & 60.3   & 72.8      &86.4        \\
%    \hline
%\end{tabular}
%\label{tab:cross}
%\end{table}

\section{Conclusion}

In this paper, we proposed a multi-stream architecture, which first uses different streams to learn different aspects of the feature maps, and then merges them together to obtain some union characteristics that can not be learned independently. Based on this architecture, we constructed a new model termed Three-Stream Convolutional Networks (TSCN), which can make full use of the learned feature maps. To further reuse the feature maps, we proposed two multi-scale architectures with the strategies of multi-scale and upsampling. We investigated the function of different streams, the methods and layers of fusion, the number of stream. Our results suggest the importance of learning correspondences between different streams. We also showed that a full connection layer in the multi-stream networks is not necessary. The comprehensive experiments indicate that our proposed networks can achieve the performance superior to the existing state-of-the-art models on iLIDS-VID, PRID-2011 and MARS datasets. As a multi-stream architecture will generate multi-vectors, how to use outer product to obtain an efficient feature representation would be an interesting
extension in further studies.

%------------------------------------------------------------------------

{\small
\bibliographystyle{ieee}
\bibliography{egbib}
}

\end{document}